\documentclass[10pt,twocolumn,letterpaper]{article}

\usepackage[pagenumbers]{cvpr} % To force page numbers, e.g. for an arXiv version

\usepackage{booktabs}       % professional-quality tables
\usepackage{multirow}       % multi-row cells in tables
\usepackage{graphicx}       % includegraphics
\usepackage{subcaption}     % subfigures
\usepackage{xcolor}         % extended colors
\usepackage{colortbl}       % table cell coloring
\usepackage{siunitx}        % number alignment in tables
\usepackage{enumitem}       % compact lists
\usepackage{pifont}         % ding symbols (e.g. \ding{55} for ✗)

\usepackage{tikz}
\usetikzlibrary{arrows.meta, calc, decorations.pathreplacing, fit}

\usepackage[most]{tcolorbox} % add this in your preamble
\newtcolorbox{promptbox}{
  colback=gray!5,    % background color
  colframe=gray!60,  % border color
  fonttitle=\bfseries,
  coltitle=black,
  boxrule=0.8pt,
  arc=3pt,           % rounded corners
  left=6pt, right=6pt, top=6pt, bottom=6pt,
  title=Prompt
}

\definecolor{cvprblue}{rgb}{0.21,0.49,0.74}
\usepackage[pagebackref,breaklinks,colorlinks,allcolors=cvprblue]{hyperref}

\def\paperID{*****} % *** Enter the Paper ID here
\def\confName{CVPR}
\def\confYear{2026}

\title{Demographic Fairness in Multimodal LLMs:\\ A Benchmark of Gender and Ethnicity Bias in Face Verification}

\author{\vspace{5pt}{\"U}nsal {\"O}zt{\"u}rk\thanks{Equal contribution} \quad Hatef Otroshi Shahreza\footnotemark[1]  \quad  S{\'e}bastien Marcel\\
Idiap Research Institute, Switzerland\\
{\small\tt \{unsal.ozturk, hatef.otroshi, sebastien.marcel\}@idiap.ch}
}
\begin{document}
\maketitle
\begin{abstract}
Multimodal Large Language Models (MLLMs) have recently been explored as face verification systems that determine whether two face images are of the same person. Unlike dedicated face recognition systems, MLLMs approach this task through visual prompting and rely on general visual and reasoning abilities. However, the demographic fairness of these models remains largely unexplored. In this paper, we present a benchmarking study that evaluates nine open-source MLLMs from six model families, ranging from 2B to 8B parameters, on the IJB-C and RFW face verification protocols across four ethnicity groups and two gender groups. We measure verification accuracy with the Equal Error Rate and True Match Rate at multiple operating points per demographic group, and we quantify demographic disparity with four FMR-based fairness metrics. Our results show that FaceLLM-8B, the only face-specialised model in our study, substantially outperforms general-purpose MLLMs on both benchmarks. The bias patterns we observe differ from those commonly reported for traditional face recognition, with different groups being most affected depending on the benchmark and the model. We also note that the most accurate models are not necessarily the fairest and that models with poor overall accuracy can appear fair simply because they produce uniformly high error rates across all demographic groups. 
% Code and annotations will be available upon acceptance.
Project page: \href{https://www.idiap.ch/paper/mllm-fairness/}{https://www.idiap.ch/paper/mllm-fairness}
\end{abstract}
\section{Introduction}
\label{sec:intro}

% Figure: demographic pairs → MLLM → similarity scores → FMR per group → fairness
%
% Self-contained tikz library loads (safe to duplicate if already in preamble)
\usetikzlibrary{arrows.meta, calc, decorations.pathreplacing, fit}

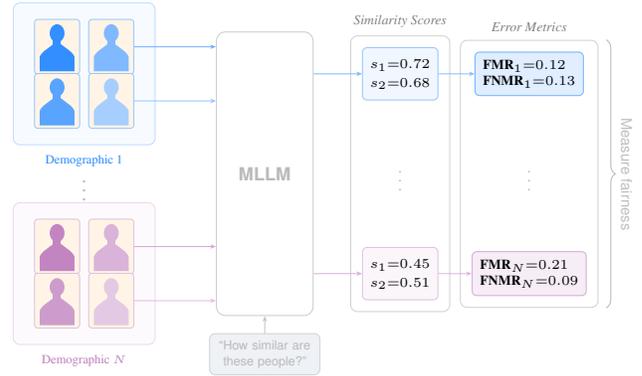
\begin{figure}[t]
  \centering
  \resizebox{\columnwidth}{!}{%
  \begin{tikzpicture}[
      every node/.style={font=\scriptsize},
      arr/.style={-{Stealth[length=2.5pt,width=2pt]}, line width=0.45pt},
      lbl/.style={font=\fontsize{5.5}{6.5}\selectfont},
      % guide.tex–style boxes
      restab/.style={draw, rounded corners=2pt, inner sep=3.5pt,
                     line width=0.4pt,
                     font=\fontsize{6}{7}\selectfont, align=left},
  ]

    % === Colours ===
    \colorlet{grpA}{blue!55!cyan}
    \colorlet{grpB}{violet!60}
    \definecolor{cardBg}{HTML}{FFF4E6}
    \definecolor{promptBg}{HTML}{EEF1F5}

    % === Bust silhouette ===
    \newcommand{\person}[2]{%
      \begin{scope}[shift={(#1.center)}]
        \fill[#2] (0, 0.21) ellipse (0.115cm and 0.14cm);
        \fill[#2] (-0.055, 0.02) rectangle (0.055, 0.10);
        \fill[#2]
            (-0.06, 0.04)
            .. controls (-0.12,-0.01) and (-0.25,-0.08) .. (-0.27,-0.16)
            -- (-0.27,-0.38) -- (0.27,-0.38) -- (0.27,-0.16)
            .. controls (0.25,-0.08) and (0.12,-0.01) .. (0.06, 0.04)
            -- cycle;
      \end{scope}%
    }

    % ================================================================
    %  STAGE 1 — Demographic groups (2 trial rows each)
    % ================================================================

    % ---------- Demographic 1 ----------
    \node[draw=grpA!25, rounded corners=3pt, fill=grpA!3,
          minimum width=2.2cm, minimum height=2.2cm,
          line width=0.35pt, inner sep=0pt]
          (dem1) at (0, 1.55) {};

    \node[draw=grpA!40, rounded corners=1.5pt, fill=cardBg,
          minimum width=0.7cm, minimum height=0.85cm,
          line width=0.3pt, inner sep=0pt] (s1a) at (-0.42, 1.97) {};
    \node[draw=grpA!40, rounded corners=1.5pt, fill=cardBg,
          minimum width=0.7cm, minimum height=0.85cm,
          line width=0.3pt, inner sep=0pt] (s1b) at ( 0.42, 1.97) {};
    \person{s1a}{grpA!80}  \person{s1b}{grpA!50}

    \node[draw=grpA!40, rounded corners=1.5pt, fill=cardBg,
          minimum width=0.7cm, minimum height=0.85cm,
          line width=0.3pt, inner sep=0pt] (s1c) at (-0.42, 1.13) {};
    \node[draw=grpA!40, rounded corners=1.5pt, fill=cardBg,
          minimum width=0.7cm, minimum height=0.85cm,
          line width=0.3pt, inner sep=0pt] (s1d) at ( 0.42, 1.13) {};
    \person{s1c}{grpA!65}  \person{s1d}{grpA!35}

    \node[lbl, text=grpA!85] at (0, 0.20) {Demographic~1};

    % ---------- vdots ----------
    \node[font=\fontsize{9}{11}\selectfont, text=black!30]
          at (0, -0.15) {$\vdots$};

    % ---------- Demographic N ----------
    \node[draw=grpB!25, rounded corners=3pt, fill=grpB!3,
          minimum width=2.2cm, minimum height=2.2cm,
          line width=0.35pt, inner sep=0pt]
          (demN) at (0, -1.55) {};

    \node[draw=grpB!40, rounded corners=1.5pt, fill=cardBg,
          minimum width=0.7cm, minimum height=0.85cm,
          line width=0.3pt, inner sep=0pt] (sNa) at (-0.42, -1.13) {};
    \node[draw=grpB!40, rounded corners=1.5pt, fill=cardBg,
          minimum width=0.7cm, minimum height=0.85cm,
          line width=0.3pt, inner sep=0pt] (sNb) at ( 0.42, -1.13) {};
    \person{sNa}{grpB!80}  \person{sNb}{grpB!50}

    \node[draw=grpB!40, rounded corners=1.5pt, fill=cardBg,
          minimum width=0.7cm, minimum height=0.85cm,
          line width=0.3pt, inner sep=0pt] (sNc) at (-0.42, -1.97) {};
    \node[draw=grpB!40, rounded corners=1.5pt, fill=cardBg,
          minimum width=0.7cm, minimum height=0.85cm,
          line width=0.3pt, inner sep=0pt] (sNd) at ( 0.42, -1.97) {};
    \person{sNc}{grpB!65}  \person{sNd}{grpB!35}

    \node[lbl, text=grpB!85] at (0, -2.90) {Demographic~$N$};

    % ================================================================
    %  STAGE 2 — MLLM box
    % ================================================================
    \node[draw=gray!40, rounded corners=4pt, fill=white,
          minimum width=1.5cm, minimum height=4.4cm,
          line width=0.45pt,
          font=\fontsize{8}{10}\selectfont\sffamily\bfseries,
          text=gray!55, inner sep=3pt]
          (M) at (2.80, 0) {MLLM};

    \node[draw=gray!35, rounded corners=2.5pt, fill=promptBg,
          font=\fontsize{5.5}{7}\selectfont\sffamily,
          text=gray!55, inner sep=3pt, align=center,
          text width=1.5cm]
          (P) at (2.80, -2.80) {``How similar are\\ these people?''};
    \draw[arr, gray!45] (P.north) -- (M.south);

    % ================================================================
    %  STAGE 3 — Similarity Scores (global box, per-group sub-boxes)
    % ================================================================

    % -- Per-group sub-boxes (colour-coded) --
    \node[restab, fill=grpA!6, draw=grpA!40]
          (scBox1) at (4.90, 1.55) {%
      \renewcommand{\arraystretch}{1.15}%
      \begin{tabular}{@{}l@{}}
        $s_1{=}0.72$ \\
        $s_2{=}0.68$ \\
      \end{tabular}};

    \node[font=\fontsize{7}{9}\selectfont, text=black!30]
          at (4.90, 0) {$\vdots$};

    \node[restab, fill=grpB!6, draw=grpB!40]
          (scBoxN) at (4.90, -1.55) {%
      \renewcommand{\arraystretch}{1.15}%
      \begin{tabular}{@{}l@{}}
        $s_1{=}0.45$ \\
        $s_2{=}0.51$ \\
      \end{tabular}};

    % -- Global enclosing box with title --
    \node[draw=gray!35, rounded corners=3pt, fill=white, fill opacity=0,
          line width=0.45pt, inner sep=5pt,
          fit=(scBox1)(scBoxN),
          label={[font=\fontsize{6}{7}\selectfont\itshape,
                  text=black!50, anchor=south]above:Similarity Scores}]
          (scGlobal) {};

    % ================================================================
    %  STAGE 4 — Per-group FMR (global box + sub-boxes)
    % ================================================================
    \node[restab, fill=grpA!12, draw=grpA!55,
          font=\fontsize{6}{7}\selectfont\bfseries]
          (fmr1) at (6.90, 1.55) {$\text{FMR}_1{=}0.12$\\$\text{FNMR}_1{=}0.13$};

    \node[font=\fontsize{7}{9}\selectfont, text=black!30]
          at (6.90, 0) {$\vdots$};

    \node[restab, fill=grpB!12, draw=grpB!55,
          font=\fontsize{6}{7}\selectfont\bfseries]
          (fmrN) at (6.90, -1.55) {$\text{FMR}_N{=}0.21$\\$\text{FNMR}_N{=}0.09$};

    % -- Global enclosing box with title --
    \node[draw=gray!35, rounded corners=3pt, fill=white, fill opacity=0,
          line width=0.45pt, inner sep=5pt,
          fit=(fmr1)(fmrN),
          label={[font=\fontsize{6}{7}\selectfont\itshape,
                  text=black!50, anchor=south]above:Error Metrics}]
          (fmrGlobal) {};

    % ================================================================
    %  Arrows (drawn last for clean layering)
    % ================================================================
    % Trial rows → MLLM (from INSIDE the demographic box)
    \draw[arr, grpA!55] (s1b.east) -- (M.west |- s1b);
    \draw[arr, grpA!40] (s1d.east) -- (M.west |- s1d);
    \draw[arr, grpB!55] (sNb.east) -- (M.west |- sNb);
    \draw[arr, grpB!40] (sNd.east) -- (M.west |- sNd);

    % MLLM → similarity score boxes
    \draw[arr, grpA!50] (M.east |- scBox1) -- (scBox1.west);
    \draw[arr, grpB!50] (M.east |- scBoxN) -- (scBoxN.west);

    % Similarity scores → FMR
    \draw[arr, grpA!45] (scBox1.east) -- (fmr1.west);
    \draw[arr, grpB!45] (scBoxN.east) -- (fmrN.west);

    % ================================================================
    %  Curly brace + "measure fairness" — AFTER the FMR column
    % ================================================================
    \draw[decorate, decoration={brace, amplitude=4pt, mirror},
          gray!50, line width=0.5pt]
         ($(fmrGlobal.south east)+(0.12, 0)$)
           -- ($(fmrGlobal.north east)+(0.12, 0)$)
         coordinate[midway] (brMid);
    \node[font=\fontsize{6.5}{8}\selectfont\sffamily, text=gray!50,
          rotate=-90, anchor=north]
         at ([xshift=15pt]brMid) {Measure fairness};

  \end{tikzpicture}}%
  \caption{Pipeline overview.
  Face pairs from each demographic group are independently prompted
  through an MLLM for pairwise verification.
  The per-pair similarity scores are aggregated into group-level
  error metrics (FMR/FNMR), which are then compared across
  demographics to assess fairness.}
  \label{fig:pictogram}
\end{figure}

Face recognition is one of the most widely deployed biometric technologies.
Modern face recognition systems typically work by mapping a face image to a fixed-length numerical representation, called an embedding, using a deep neural network trained on large collections of labelled face images~\cite{facenet}.
To determine whether two images depict the same person, the system computes the distance (or similarity) between their embeddings and compares it to a decision threshold.
This embedding-based approach has reached high levels of accuracy on standard benchmarks~\cite{cosface,arcface,magface,adaface}, and it is now used in applications ranging from border control and law enforcement to mobile device authentication.

In recent years, Large Language Models (LLMs) have transformed natural language processing by demonstrating strong performance on a broad range of tasks through a single, general-purpose architecture~\cite{llama, google-emergent}.
This success has led to the development of Multimodal Large Language Models (MLLMs), which extend LLMs with the ability to process visual inputs alongside text~\cite{mllm-survey,llava,team2023gemini,hurst2024gpt,claude35}.
Models such as Qwen-VL~\cite{qwen2025qwen25technicalreport}, LLaVA~\cite{llava,liu2023improvedllava}, and Idefics~\cite{laurencon2024building} can accept one or more images together with a text prompt, and produce a free-form textual response.
This capability opens up the possibility of using MLLMs for face verification: given two face images, the model can be prompted to judge whether they belong to the same person, and its response can be converted into a similarity score.
Unlike embedding-based systems, which require specialised training on face identity labels, MLLMs approach the task through visual question answering by relying on the general visual and reasoning abilities acquired during pretraining.
Whether this approach can produce accurate and reliable verification scores is an open question.

A separate but equally important concern is the fairness of face recognition systems across different demographic groups~\cite{drozdowski-bias-survey, ketan-fairness-survey}.
Face recognition systems have been shown to perform unevenly across demographic groups defined by attributes such as ethnicity and gender, and studies have repeatedly found that certain groups tend to have higher error rates than others.~\cite{jjhoward-1}.
This raises serious concerns when face recognition is used in high-stakes applications, because errors can have direct consequences for individuals~\cite{frvt-3}.
Fairness in this context means that the system should produce similar error rates across demographic groups at an operating point; if one group has a substantially higher false match rate or false non-match rate than another, the system is said to be biased against that group.~\cite{frvt-8}.

While demographic fairness has been studied extensively for embedding-based face recognition, no comparable analysis exists for MLLM-based face verification.
It is not obvious that the bias patterns observed in traditional systems will carry over to MLLMs, since the two approaches differ fundamentally in how they process face images and arrive at a similarity judgment.
Understanding the fairness properties of MLLMs is important, because these models are increasingly being considered for tasks that were previously handled by specialised systems.

In this paper, we present a systematic demographic fairness evaluation of MLLMs for face verification.
We evaluate nine MLLMs on the IJB-C and RFW verification protocols across four ethnicity groups and two gender groups.
We measure verification accuracy with the Equal Error Rate (EER) and True Match Rate (TMR) at standard operating points, and we quantify demographic disparity with four complementary FMR-based fairness metrics at multiple thresholds.
Our results show that no current MLLM approaches the accuracy of dedicated embedding-based systems, and that the bias patterns in MLLMs differ from those observed in traditional face recognition.

\section{Related Work}
\label{sec:related}

\subsection{Bias in Face Recognition}
Face recognition systems that use a single global decision threshold can produce different error rates for different demographic groups. Buolamwini and Gebru~\cite{gendershades} demonstrated this for commercial gender classifications systems and concluded that darker-skinned female faces had higher error rates. Similarly, Howard et al.~\cite{jjhoward-1} showed that score distributions of different demographic groups are different and therefore have an impact on system performance across demographic groups. Further studies examined how facial attributes such as hairstyle~\cite{hairy} and demographic-dependent properties of face regions~\cite{face-regions} contribute to performance gaps between different groups. Sarridis et al.~\cite{demographic-bias-analysis} similarly analysed intersectional biases over race, age, and gender and concluded that certain subgroups such as older African females experience larger error rates. Kotwal and Marcel~\cite{ketan-fairness-survey} provide a comprehensive survey of demographic fairness in face recognition.

\subsection{Fairness Metrics for Biometric Systems}
Following the evidence of demographic bias in biometric systems, Grother et al.~\cite{frvt-3} at NIST conducted a large-scale evaluation of over 100 commercially available face recognition systems as part of the Face Recognition Vendor Test, and reported FMR and FNMR values for various demographic groups. This prompted work on metrics that reduce per-group error rates into a scalar summarizing the fairness of a biometric system as a combination of values measuring the ``spread'' of in-group FMR and FNMR values obtained by setting a global threshold. De Freitas Pereira and Marcel~\cite{fdr} proposed the Fairness Discrepancy Rate (FDR), which is based on the maximum differential of FMR and FNMR across groups. A subsequent NIST report by Grother~\cite{frvt-8} confirmed that large demographic differentials persist across algorithms and introduced the Inequity Rate (IR), based on linear combinations of maximum FMR and FNMR ratios, and the Worst-case Error-Rate Metric (WERM), which takes the exponent-weighted product of the maximum FMR and FNMR ratios to the geometric mean across all groups. Howard et al.~\cite{garbe} proposed the Gini Aggregation Rate for Biometric Equitability (GARBE), which uses the Gini index of per-group FMR and FNMR values and showed that it better satisfies a set of interpretability criteria than previous measures. Kotwal and Marcel~\cite{ketan-fairness-distribution} further proposed fairness measures based on pre-decision score distributions rather than post-decision error rates. These efforts were eventually consolidated into ISO/IEC~19795-10:2024~\cite{iso19795-10:2024}, which standardises the reporting of biometric performance variation across demographic groups. The standard specifies that FMR and FNMR differentials and ratios should be reported per group, but it does not prescribe a single scalar fairness metric. In this paper, we mainly use metrics equivalent to FMR-based metrics described in the ISO standard.

\subsection{Face recognition with MLLMs}

Several papers have explored the applications of MLLMs for face-related tasks, including multimodal reasoning, face recognition, attribute analysis, deepfake detection, and anti-spoofing. A recent survey \cite{shahreza2025foundation} provides a comprehensive overview of the applications of MLLMs and foundation models in biometrics and face recognition.

Early studies explored the use of pretrained MLLMs, such as ChatGPT~\cite{hurst2024gpt}, for face verification and predicting soft biometrics, such as age, gender, and ethnicity~\cite{hassanpour2024chatgpt}. 
Jia \textit{et al.} \cite{jia2024can} also used ChatGPT for zero-shot face deepfake detection.  
Shi \textit{et al.} \cite{shi2024shield} explored chain-of-thoughts prompting for  ChatGPT~\cite{hurst2024gpt} and Gemini~\cite{team2023gemini}  for face anti-spoofing and  deepfake detection. 
Komaty \textit{et al.} ~\cite{komaty2025exploring} explored in-context learning with ChatGPT~\cite{hurst2024gpt} for face anti-spoofing. 
Wang \textit{et al.} proposed FaceBench~\cite{wang2025facebench} as a visual question-answering benchmark for facial attributes. Narayan \textit{et al.}  proposed FaceXBench~\cite{narayan2025facexbench}  to benchmark MLLMs on different face-related tasks, including expression recognition, attribute prediction,  anti-spoofing, etc.  FaceRecBench~\cite{facerecbench2025} was also proposed for benchmarking MLLMs for face verification. 
FaceXBench~\cite{narayan2025facexbench} used multiple-choice questions  and FaceRecBench~\cite{facerecbench2025} used Yes/No questions for evaluating the performance of MLLMs for face verification. 
In \cite{shahreza2026evaluating}, MLLMs were benchmarked for heterogeneous face recognition.
In this paper, we focus on fairness of MLLMs for face verification and benchmark bias MLLMs with multiple fairness metrics, which has not been studied for MLLMs.
% Compared to previous benchmarks, we use similarity scores from MLLM and report detailed analyses at different thresholds.

\section{Methodology}
\label{sec:method}
This section describes our benchmarking methodology. \Cref{fig:pictogram} summarises our pipeline.

\subsection{Face Verification with MLLMs}
\label{sec:prompting}
To evaluate MLLMs for face verification, we provide MLLM with two face images and a text prompt. In the text prompt, we ask the model to compare the given images and return a similarity score:
\begin{promptbox}
"On a scale from 0 to 100, how likely (as a single number) are these two face images of the same person? Only output a single number (no other text)."
\end{promptbox}
\noindent We use the output of MLLM as similarity score (normalised to $[0,1]$.) to evaluate the model for face verification.

\subsection{Datasets and Protocol}
\label{sec:dataset}
We evaluate on two face verification benchmarks.

\textbf{IJB-C.}
The IARPA Janus Benchmark-C (IJB-C)~\cite{ijbc} contains 31{,}334 still images and 117{,}542 video frames of 3{,}531 subjects. As per the protocol, the images are organised into \emph{biometric templates}, where each template can contain anywhere from one image to several hundred frames. The 1:1 verification protocol defines 19{,}557 template pairs, roughly half of which are genuine (same identity) and half impostor (different identities). We labeled IJB-C with gender (\textbf{Male}, \textbf{Female}) and ethnicity labels (\textbf{African}, \textbf{East Asian}, \textbf{South Asian}, \textbf{Caucasian}). Due to the size of the protocol, we subsample 10{,}000 template comparisons, which results in 7{,}200 to 9{,}990 scored pairs per model depending on the model's ability to produce a valid score for each pair. As a result of this sampling, we get 2{,}750 cross-ethnicity comparisons; and 670, 800, 700, 5{,}000, within-ethnicity comparisons for Africans, East Asians, South Asians, and Caucasians respectively. As for gender, we get 3{,}000 and 4{,}500 in-group female and male comparisons, and 2{,}500 cross-group impostor comparisons. This subset of IJB-C, took on average about 20 days of GPU compute on an NVIDIA H100 per MLLM to be evaluated, and scaling to the entire benchmark remains challenging.

\textbf{RFW.}
Racial Faces in-the-Wild (RFW)~\cite{rfw} is a balanced face verification benchmark designed specifically for studying ethnic bias. It contains four ethnicity groups (African, Asian, Caucasian, Indian), and each group contributes exactly 3{,}000 genuine and 3{,}000 impostor pairs, for a total of 24{,}000 pairs. RFW does not carry gender annotations, and we therefore do not consider this dataset for measuring gender-fairness.

\subsection{Evaluated Models}
\label{sec:models}

We evaluate nine open-source and publicly available MLLMs from six model families (\cref{tab:models}).
Eight of these, including Idefics3-8B-Llama3~\cite{laurencon2024building}, LLaVA-NeXT-Mistral-7B~\cite{llava}, Ovis1.5-Llama3-8B~\cite{lu2024ovis}, Qwen2-VL-2B~\cite{Qwen2VL}, Qwen2-VL-7B~\cite{Qwen2VL}, Qwen2.5-VL-3B~\cite{qwen2025qwen25technicalreport}, Qwen2.5-VL-7B~\cite{qwen2025qwen25technicalreport}, and Valley2~\cite{wu2025valley2}, are general-purpose vision-language models that were not specifically designed for face-related tasks.
The ninth, FaceLLM-8B~\cite{shahreza2025facellm}, is fine-tuned for face analysis and is therefore the only model in our study with specialised face knowledge.
All evaluations are performed in a zero-shot setting, meaning that the models receive no fine-tuning and no in-context examples for the verification task.

\subsection{Template-Level Scoring}
\label{sec:scoring}

Since templates can contain multiple images or video frames, we need a way to reduce a template pair to a single similarity score.
Given a probe template $T_p = \{x_1^p, \dots, x_m^p\}$ and a gallery template $T_g = \{x_1^g, \dots, x_n^g\}$, we form all $m \times n$ image pairs.
For each pair $(x_i^p, x_j^g)$, we prompt the MLLM in a zero-shot fashion with both face images and a standardised textual query that asks whether the two images depict the same person, for which the model returns a similarity score $s_{ij} \in [0,1]$. The final template-level score is then computed as the arithmetic mean over all pairwise scores:
\begin{equation}
  s(T_p, T_g) = \frac{1}{mn} \sum_{i=1}^{m} \sum_{j=1}^{n} s_{ij}.
  \label{eq:template_score}
\end{equation}
One advantage of this approach is that templates with many images benefit from noise reduction: individual pairwise scores may be noisy, but averaging over many pairs produces a more stable estimate. This comes at the cost of quadratically many inferences.

\subsection{Verification Metrics}
\label{sec:ver_metrics}

To evaluate verification performance, we sweep a decision threshold $\tau$ across the range $[0,1]$ in steps of $0.005$ (201 operating points) and compute the following error rates at each threshold:
\begin{itemize}[nosep,leftmargin=*]
    \item \textbf{False Match Rate} (FMR): the proportion of impostor pairs whose similarity score meets or exceeds the threshold~$\tau$. A high FMR means the system incorrectly accepts many impostor pairs as genuine.
    \item \textbf{False Non-Match Rate} (FNMR): the proportion of genuine pairs whose similarity score falls below~$\tau$. A high FNMR means the system incorrectly rejects many genuine pairs.
    \item \textbf{Equal Error Rate} (EER): the operating point at which the FMR and FNMR are equal. The EER is a single number that summarises the overall accuracy of the system.
\end{itemize}
All metrics are computed both globally (over all pairs) and separately for each demographic group. We also report the True Match Rate ($\text{TMR} = 1 - \text{FNMR}$) at fixed FMR thresholds of 10\%, 1\%, and 0.1\%, which correspond to increasingly strict security requirements. The full operating characteristic is plotted in~\cref{fig:det}.

%\input{fig/det_all_models}
% Auto-generated by generate_tables_unified.py
\begin{table}[t]
  \caption{Evaluated MLLMs.  \emph{Params} denotes approximate total parameter count.}
  \label{tab:models}
  \centering
  \small
  \begin{tabular}{@{}llr@{}}
    \toprule
    Model & Family & Params \\
    \midrule
    Idefics3-8B-Llama3~\cite{laurencon2024building}        & Idefics3     & 8B \\
    Ovis1.5-Llama3-8B~\cite{lu2024ovis}         & Ovis         & 8B \\
    Qwen2-VL-2B~\cite{Qwen2VL}               & Qwen2-VL     & 2B \\
    Qwen2-VL-7B~\cite{Qwen2VL}               & Qwen2-VL     & 7B \\
    Qwen2.5-VL-3B~\cite{qwen2025qwen25technicalreport}             & Qwen2.5-VL   & 3B \\
    Qwen2.5-VL-7B~\cite{qwen2025qwen25technicalreport}             & Qwen2.5-VL   & 7B \\
    Valley2~\cite{wu2025valley2}                   & Valley       & 7B \\
    LLaVA-NeXT-Mistral-7B~\cite{llava}    & LLaVA-NeXT   & 7B \\
    \midrule
    FaceLLM-8B~\cite{shahreza2025facellm}                & FaceLLM      & 8B \\
    \bottomrule
  \end{tabular}
\end{table}

\subsection{Fairness Metrics}
\label{sec:fair_metrics}

We compute four fairness metrics based on per-group FMRs and evaluate these metrics at the global EER threshold as well as at three fixed FMR operating points (10\%, 1\%, 0.1\%).
Let $\mathcal{G} = \{g_1, \dots, g_K\}$ denote the $K$ demographic groups, and let $\text{FMR}_g$ be the false match rate for group $g$ at a given decision threshold.
For brevity, we write $e_g \equiv \text{FMR}_g$ and use the subscript~F throughout (e.g.\ $\Delta_\text{F}$, $R_\text{F}$, $M_\text{F}$, $G_\text{F}$).

\textbf{Maximum FMR differential.}
The differential measures the range between the highest and lowest group error rates:
\begin{equation}
  \Delta = \max_{g}\, e_g - \min_{g}\, e_g
  \label{eq:diff}
\end{equation}
Since this is an absolute measure, its magnitude depends on the overall level of FMR: a system with high FMR can have a large $\Delta$ even if the relative differences between groups are small. Perfect fairness achieved at $\Delta=0$.

\textbf{Maximum FMR ratio.}
The FMR ratio compares the worst-case group to the best-case group:
\begin{equation}
  R = \frac{\max_{g}\, e_g}{\min_{g}\, e_g}
  \label{eq:ratio}
\end{equation}
A value of one indicates that all groups have the same FMR.
The ratio is a relative measure, which makes it useful for comparing disparity across different operating points.
However, it becomes unstable when the minimum per-group FMR approaches zero, because even a small absolute difference in the numerator leads to a very large ratio.

\textbf{Maximum ratio to geometric mean.}
The maximum ratio to geometric mean normalises the worst-case group FMR by the geometric mean of all group FMR values:
\begin{equation}
  M = \frac{\max_g\, e_g}{ \bigl(\prod_g e_g\bigr)^{1/K}},
  \label{eq:multi}
\end{equation}
A value of $M = 1$ means that the worst-case group has the same FMR as the group average.
Compared to the ratio $R$, which divides by the minimum group FMR, the multi-group metric is more stable because the geometric mean pools information from all groups and is less sensitive to a single group with a very low FMR.

\textbf{Gini coefficient.}
The Gini coefficient indicates the inequality across all groups:
\begin{equation}
  G = \frac{\sum_{i=1}^{K}\sum_{j=1}^{K} |e_i - e_j|}{2K\sum_{i=1}^{K} e_i}.
  \label{eq:gini}
\end{equation}
A value of $G = 0$ means that all groups have the same FMR, As $G$ approaches $1$, the FMR concentrates in a single group while the rest of the groups have near-zero FMR.

\textbf{``Decidability'' index.}
In addition to threshold-dependent metrics, we compute the decidability index ($d'$) for each demographic group as defined in~\cite{decidability}:
\begin{equation}
  d' = \frac{\mu_{\text{genuine}} - \mu_{\text{impostor}}}{\sqrt{\tfrac{1}{2}\bigl(\sigma_{\text{genuine}}^{2}+\sigma_{\text{impostor}}^{2}\bigr)}},
  \label{eq:dprime}
\end{equation}
where $\mu_{.}$ and $\sigma_{.}$ are the mean and standard deviation of genuine and impostor score distributions. This measure quantifies how well the genuine and impostor score distributions are separated, independently of any particular threshold choice and can be interpreted as the number of pooled standard deviations that separate the two distributions.
\section{Experimental Results}
\label{sec:results}

% Figure: DET curves — all models on one plot (log-log scale)
\begin{figure}[t]
  \centering
  \scalebox{0.90} {
  \includegraphics[width=\linewidth]{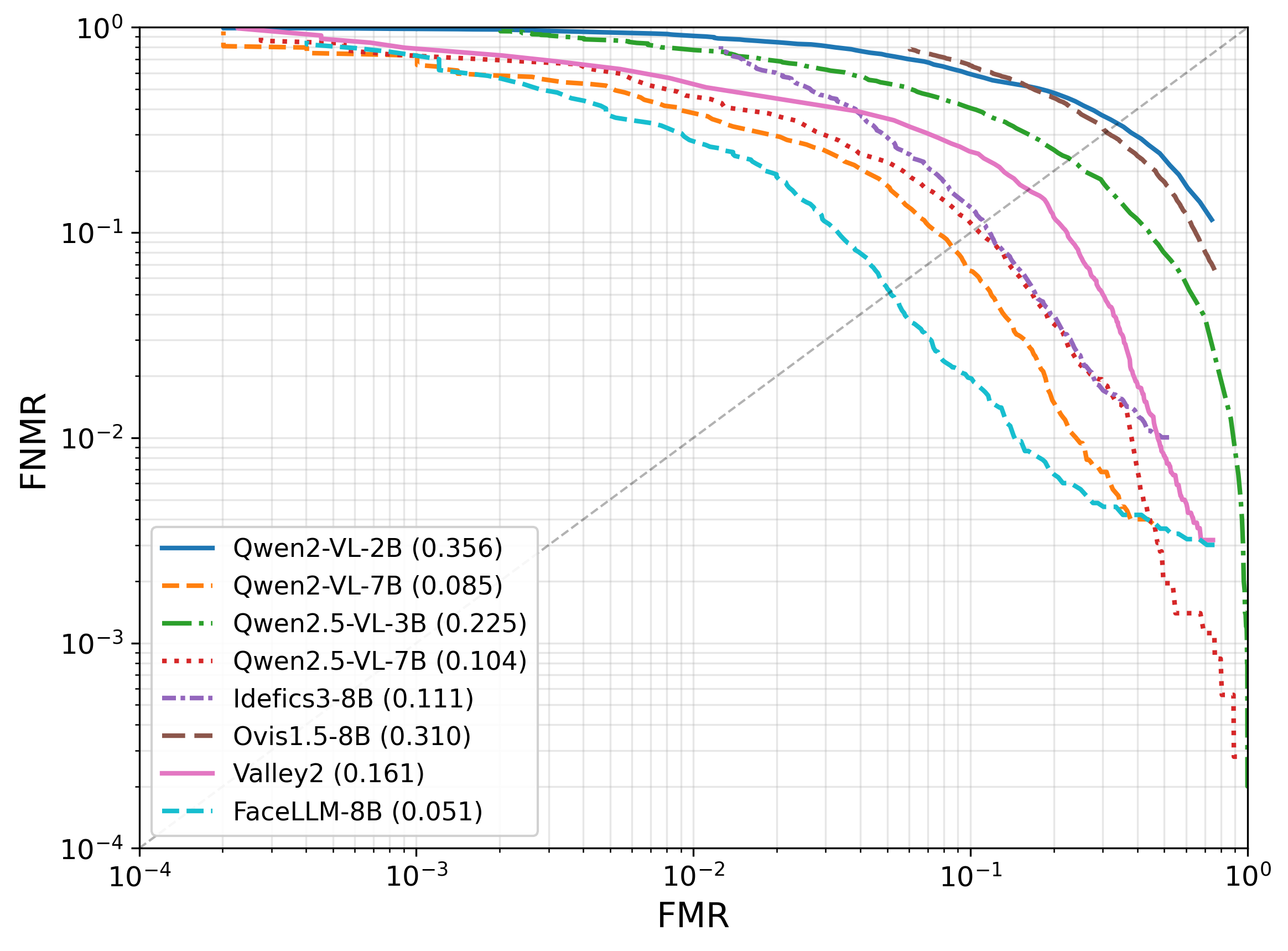}
  }
  \caption{Detection Error Trade-off (DET) curves for all evaluated models on the IJB-C 1:1 verification protocol.  Lower-left is better.}
  \label{fig:det}
\end{figure}

\Cref{tab:models} lists the nine MLLMs evaluated in this study.
LLaVA-NeXT-Mistral-7B produces only two unique similarity scores and fails to provide meaningful verification performance. We exclude it from all subsequent analyses.

\subsection{Verification Performance}
\label{sec:overall_perf}

% Auto-generated by generate_tables_unified.py
\begin{table*}[!htbp]
  \caption{Verification performance on IJB-C and RFW.  EER~(\%, $\downarrow$): Equal Error Rate; per-group EER is at each group's own operating point.  $\sigma$: standard deviation of per-ethnicity EER.  Gender EER: female and male EER with absolute gap~$\Delta$ ($\downarrow$ is more uniform); not available for RFW (marked {---}).  TMR~(\%, $\uparrow$): True Match Rate at fixed FMR.  RFW uses ``Asian'' and ``Indian'' for the East-Asian and South-Asian groups, respectively.  Best per column within each block is \textbf{bolded}.  {\ding{55}}~indicates an unreachable operating point.}
  \label{tab:performance}
  \centering
  \footnotesize
  \setlength{\tabcolsep}{3pt}
  \scalebox{0.97}{
  \begin{tabular}{@{}l S[table-format=2.2] S[table-format=2.2] S[table-format=2.2] S[table-format=2.2] S[table-format=2.2] S[table-format=1.2] S[table-format=2.2] S[table-format=2.2] S[table-format=1.2] S[table-format=2.2] S[table-format=2.2] S[table-format=2.2]@{}}
    \toprule
    & \multicolumn{6}{c}{Ethnicity EER (\%, $\downarrow$)} & \multicolumn{3}{c}{Gender EER (\%, $\downarrow$)} & \multicolumn{3}{c}{TMR (\%, $\uparrow$)} \\
    \cmidrule(lr){2-7} \cmidrule(lr){8-10} \cmidrule(lr){11-13}
    Model & {Global} & {African} & {Cauc.} & {E.\,As.} & {S.\,As.} & {$\sigma$} & {Female} & {Male} & {$\Delta$} & {@10\%} & {@1\%} & {@0.1\%} \\
    \midrule
    \rowcolor{green!12}
    \multicolumn{13}{c}{\textbf{IJB-C}} \\
    Idefics3-8B-Llama3        & 11.21 & 8.79 & 11.90 & 10.72 & 8.85 & 1.31 & 12.46 & 9.98 & 2.48 & 84.39 & \ding{55} & \ding{55} \\
    Ovis1.5-Llama3-8B         & 30.89 & 32.71 & 30.35 & 31.30 & 32.67 & 0.99 & 30.18 & 31.65 & 1.46 & 34.59 & \ding{55} & \ding{55} \\
    Qwen2-VL-2B               & 33.69 & 29.75 & 35.78 & 34.00 & 29.36 & 2.75 & 35.63 & 31.86 & 3.77 & 40.26 & 8.46 & 1.56 \\
    Qwen2-VL-7B               & 8.54 & 6.34 & 9.14 & 7.49 & 5.97 & 1.23 & 8.28 & 8.17 & \textbf{0.10} & 93.42 & 60.38 & 25.87 \\
    Qwen2.5-VL-3B             & 22.89 & 23.59 & 22.18 & 22.86 & 23.46 & \textbf{0.56} & 23.17 & 22.05 & 1.12 & 59.07 & 21.90 & \ding{55} \\
    Qwen2.5-VL-7B             & 10.43 & 9.24 & 11.07 & 10.38 & 8.13 & 1.12 & 10.99 & 9.38 & 1.61 & 88.82 & 53.45 & \textbf{26.39} \\
    Valley2                   & 16.17 & 16.03 & 15.26 & 19.91 & 13.99 & 2.21 & 16.83 & 15.00 & 1.83 & 75.11 & 43.33 & 20.46 \\
    \midrule
    FaceLLM-8B                & \textbf{5.13} & \textbf{4.65} & \textbf{5.53} & \textbf{4.31} & \textbf{3.98} & 0.58 & \textbf{5.37} & \textbf{4.82} & 0.55 & \textbf{98.03} & \textbf{71.75} & 24.04 \\
    \midrule
    \rowcolor{cyan!10}
    \multicolumn{13}{c}{\textbf{RFW}} \\
    Idefics3-8B-Llama3        & 35.08 & 41.95 & 34.47 & 29.02 & 34.63 & 4.60 & {---} & {---} & {---} & \ding{55} & \ding{55} & \ding{55} \\
    Ovis1.5-Llama3-8B         & 49.18 & 52.83 & 49.17 & 45.13 & 49.58 & 2.73 & {---} & {---} & {---} & \ding{55} & \ding{55} & \ding{55} \\
    Qwen2-VL-2B               & 50.35 & 53.32 & 50.15 & 50.02 & 48.43 & \textbf{1.77} & {---} & {---} & {---} & 4.21 & \ding{55} & \ding{55} \\
    Qwen2-VL-7B               & 34.39 & 39.63 & 34.57 & 24.40 & 34.13 & 5.51 & {---} & {---} & {---} & 12.98 & 0.52 & 0.52 \\
    Qwen2.5-VL-3B             & 43.14 & 48.97 & 41.73 & 38.12 & 42.63 & 3.91 & {---} & {---} & {---} & \ding{55} & \ding{55} & \ding{55} \\
    Qwen2.5-VL-7B             & 34.98 & 40.85 & 34.45 & 26.30 & 32.58 & 5.19 & {---} & {---} & {---} & 11.30 & \ding{55} & \ding{55} \\
    Valley2                   & 39.67 & 45.85 & 40.35 & 30.33 & 39.92 & 5.58 & {---} & {---} & {---} & 26.99 & 0.08 & 0.08 \\
    \midrule
    FaceLLM-8B                & \textbf{29.46} & \textbf{35.25} & \textbf{27.87} & \textbf{21.23} & \textbf{29.13} & 4.98 & {---} & {---} & {---} & \textbf{38.69} & \textbf{1.42} & \textbf{1.42} \\
    \bottomrule
  \end{tabular}
  }
\end{table*}

% Auto-generated by generate_tables_unified.py
\begin{table*}[!htbp]
  \caption{FMR-based fairness metrics at the global EER threshold and three standard operating points, plus decidability index~$d'$.  $\Delta_\text{F}$: max per-group FMR difference ($\downarrow$); $R_\text{F}$: ratio of max to min per-group FMR ($\downarrow$); $M_\text{F}$: ratio of max per-group FMR to the geometric mean ($\downarrow$); $G_\text{F}$: Gini coefficient of per-group FMR ($\downarrow$); $d'$: mean decidability index across groups ($\uparrow$).  {\ding{55}}~indicates an unreachable operating point; {$\infty$}~marks ratios where at least one group has near-zero FMR.  Best per block is \textbf{bolded}.}
  \label{tab:fairness}
  \centering
  \footnotesize
  \setlength{\tabcolsep}{2pt}
  \scalebox{0.97}{
  \begin{tabular}{@{}l S[table-format=1.3] S[table-format=2.2] S[table-format=1.2] S[table-format=1.3] S[table-format=1.3] S[table-format=2.2] S[table-format=1.2] S[table-format=1.3] S[table-format=1.3] S[table-format=2.2] S[table-format=1.2] S[table-format=1.3] S[table-format=1.3] S[table-format=2.2] S[table-format=1.2] S[table-format=1.3] S[table-format=1.2]@{}}
    \toprule
    & \multicolumn{4}{c}{@EER} & \multicolumn{4}{c}{@FMR\,=\,10\%} & \multicolumn{4}{c}{@FMR\,=\,1\%} & \multicolumn{4}{c}{@FMR\,=\,0.1\%} & \\
    \cmidrule(lr){2-5} \cmidrule(lr){6-9} \cmidrule(lr){10-13} \cmidrule(lr){14-17} \cmidrule(lr){18-18}
    Model & {$\Delta_\text{F}$} & {$R_\text{F}$} & {$M_\text{F}$} & {$G_\text{F}$} & {$\Delta_\text{F}$} & {$R_\text{F}$} & {$M_\text{F}$} & {$G_\text{F}$} & {$\Delta_\text{F}$} & {$R_\text{F}$} & {$M_\text{F}$} & {$G_\text{F}$} & {$\Delta_\text{F}$} & {$R_\text{F}$} & {$M_\text{F}$} & {$G_\text{F}$} & {$d'$} \\
    \midrule
    \rowcolor{green!12}
    \multicolumn{18}{c}{\textbf{IJB-C -- Ethnicity}} \\
    Idefics3-8B-Llama3        & 0.033 & 1.40 & \textbf{1.12} & 0.064 & \textbf{0.026} & \textbf{1.38} & \textbf{1.15} & 0.070 & \ding{55} & \ding{55} & \ding{55} & \ding{55} & \ding{55} & \ding{55} & \ding{55} & \ding{55} & 2.86 \\
    Ovis1.5-Llama3-8B         & 0.143 & 1.70 & 1.19 & 0.090 & 0.072 & 2.32 & 1.35 & 0.148 & \ding{55} & \ding{55} & \ding{55} & \ding{55} & \ding{55} & \ding{55} & \ding{55} & \ding{55} & 0.90 \\
    Qwen2-VL-2B               & 0.103 & \textbf{1.35} & 1.14 & \textbf{0.059} & 0.035 & 1.41 & 1.22 & \textbf{0.068} & 0.007 & 3.08 & 1.51 & 0.181 & 0.002 & {$\infty$} & {$\infty$} & \textbf{0.387} & 0.90 \\
    Qwen2-VL-7B               & 0.057 & 2.38 & 1.45 & 0.167 & 0.065 & 2.35 & 1.46 & 0.170 & 0.006 & 2.30 & 1.68 & 0.182 & \textbf{0.001} & {$\infty$} & {$\infty$} & 0.750 & 3.59 \\
    Qwen2.5-VL-3B             & 0.080 & 1.47 & 1.15 & 0.069 & 0.051 & 1.81 & 1.27 & 0.113 & \textbf{0.005} & \textbf{1.68} & \textbf{1.39} & \textbf{0.108} & \ding{55} & \ding{55} & \ding{55} & \ding{55} & 1.41 \\
    Qwen2.5-VL-7B             & 0.051 & 1.81 & 1.30 & 0.133 & 0.048 & 1.75 & 1.33 & 0.134 & 0.012 & 3.69 & 1.80 & 0.239 & 0.002 & {$\infty$} & {$\infty$} & 0.600 & 2.46 \\
    Valley2                   & 0.102 & 2.07 & 1.28 & 0.130 & 0.085 & 2.92 & 1.46 & 0.172 & 0.012 & {$\infty$} & {$\infty$} & 0.466 & 0.001 & {$\infty$} & {$\infty$} & 0.750 & 2.03 \\
    \midrule
    FaceLLM-8B                & \textbf{0.020} & 1.53 & 1.33 & 0.092 & 0.042 & 1.61 & 1.31 & 0.099 & 0.012 & 4.91 & 2.16 & 0.311 & 0.002 & {$\infty$} & {$\infty$} & 0.387 & \textbf{4.50} \\
    \midrule
    \rowcolor{green!12}
    \multicolumn{18}{c}{\textbf{IJB-C -- Gender}} \\
    Idefics3-8B-Llama3        & 0.041 & 1.44 & 1.20 & 0.089 & 0.035 & 1.50 & 1.22 & 0.100 & \ding{55} & \ding{55} & \ding{55} & \ding{55} & \ding{55} & \ding{55} & \ding{55} & \ding{55} & 2.70 \\
    Ovis1.5-Llama3-8B         & 0.062 & 1.22 & 1.11 & 0.051 & \textbf{0.006} & \textbf{1.06} & \textbf{1.03} & \textbf{0.015} & \ding{55} & \ding{55} & \ding{55} & \ding{55} & \ding{55} & \ding{55} & \ding{55} & \ding{55} & 0.93 \\
    Qwen2-VL-2B               & 0.088 & 1.29 & 1.14 & 0.063 & 0.026 & 1.32 & 1.15 & 0.069 & \textbf{0.000} & \textbf{1.05} & \textbf{1.02} & \textbf{0.012} & \textbf{0.000} & \textbf{1.43} & \textbf{1.20} & \textbf{0.089} & 0.82 \\
    Qwen2-VL-7B               & 0.018 & 1.25 & 1.12 & 0.055 & 0.025 & 1.31 & 1.14 & 0.067 & 0.005 & 1.86 & 1.36 & 0.151 & 0.001 & {$\infty$} & {$\infty$} & 0.500 & 3.39 \\
    Qwen2.5-VL-3B             & 0.054 & 1.28 & 1.13 & 0.061 & 0.027 & 1.33 & 1.15 & 0.071 & 0.004 & 1.49 & 1.22 & 0.098 & \ding{55} & \ding{55} & \ding{55} & \ding{55} & 1.45 \\
    Qwen2.5-VL-7B             & 0.046 & 1.60 & 1.27 & 0.116 & 0.047 & 1.65 & 1.29 & 0.123 & 0.002 & 1.29 & 1.14 & 0.063 & 0.001 & {$\infty$} & {$\infty$} & 0.500 & 2.38 \\
    Valley2                   & 0.034 & 1.24 & 1.12 & 0.055 & 0.024 & 1.28 & 1.13 & 0.061 & 0.005 & 1.84 & 1.36 & 0.147 & 0.001 & 2.12 & 1.46 & 0.179 & 2.06 \\
    \midrule
    FaceLLM-8B                & \textbf{0.001} & \textbf{1.02} & \textbf{1.01} & \textbf{0.004} & 0.007 & 1.07 & 1.03 & 0.017 & 0.004 & 1.54 & 1.24 & 0.106 & 0.001 & 2.10 & 1.45 & 0.177 & \textbf{4.31} \\
    \midrule
    \rowcolor{cyan!10}
    \multicolumn{18}{c}{\textbf{RFW -- Ethnicity}} \\
    Idefics3-8B-Llama3        & 0.399 & 2.46 & 1.58 & 0.181 & \ding{55} & \ding{55} & \ding{55} & \ding{55} & \ding{55} & \ding{55} & \ding{55} & \ding{55} & \ding{55} & \ding{55} & \ding{55} & \ding{55} & 0.74 \\
    Ovis1.5-Llama3-8B         & 0.073 & \textbf{1.10} & \textbf{1.04} & \textbf{0.020} & \ding{55} & \ding{55} & \ding{55} & \ding{55} & \ding{55} & \ding{55} & \ding{55} & \ding{55} & \ding{55} & \ding{55} & \ding{55} & \ding{55} & 0.05 \\
    Qwen2-VL-2B               & 0.240 & 1.52 & 1.16 & 0.080 & 0.112 & 12.55 & 5.75 & 0.533 & \ding{55} & \ding{55} & \ding{55} & \ding{55} & \ding{55} & \ding{55} & \ding{55} & \ding{55} & 0.02 \\
    Qwen2-VL-7B               & 0.168 & 2.98 & 1.51 & 0.176 & \textbf{0.023} & 2.41 & 1.79 & 0.191 & 0.001 & {$\infty$} & {$\infty$} & 0.625 & 0.001 & {$\infty$} & {$\infty$} & 0.625 & 0.93 \\
    Qwen2.5-VL-3B             & 0.270 & 1.84 & 1.47 & 0.131 & \ding{55} & \ding{55} & \ding{55} & \ding{55} & \ding{55} & \ding{55} & \ding{55} & \ding{55} & \ding{55} & \ding{55} & \ding{55} & \ding{55} & 0.40 \\
    Qwen2.5-VL-7B             & 0.406 & 2.97 & 1.68 & 0.205 & 0.038 & 9.21 & 2.84 & 0.395 & \ding{55} & \ding{55} & \ding{55} & \ding{55} & \ding{55} & \ding{55} & \ding{55} & \ding{55} & 0.80 \\
    Valley2                   & \textbf{0.057} & 2.02 & 1.32 & 0.121 & 0.057 & \textbf{2.02} & \textbf{1.32} & \textbf{0.121} & \textbf{0.000} & {$\infty$} & {$\infty$} & \textbf{0.500} & \textbf{0.000} & {$\infty$} & {$\infty$} & \textbf{0.500} & 0.63 \\
    \midrule
    FaceLLM-8B                & 0.478 & 4.76 & 2.31 & 0.293 & 0.240 & 9.48 & 4.16 & 0.469 & 0.002 & {$\infty$} & {$\infty$} & 0.750 & 0.002 & {$\infty$} & {$\infty$} & 0.750 & \textbf{1.08} \\
    \bottomrule
  \end{tabular}
  }
\end{table*}

\Cref{tab:performance} reports global and per-group EER together with TMR at three fixed FMR thresholds for both benchmarks.
The DET curves in \cref{fig:det} visualise the full operating characteristic for every model.

\textbf{IJB-C.} FaceLLM-8B achieves the lowest global EER (5.13\%), roughly half that of the next-best model, Qwen2-VL-7B (8.54\%).
Qwen2.5-VL-7B follows at 10.43\%, and Idefics3-8B at 11.21\%.
The remaining four general-purpose models have considerably higher error rates, with Valley2 at 16.17\%, Qwen2.5-VL-3B at 22.89\%, and Ovis1.5 and Qwen2-VL-2B near chance level (30.89\% and 33.69\%, respectively). The TMR values in \cref{tab:performance} are consistent with this ranking. At FMR\,=\,10\%, FaceLLM-8B correctly matches 98.0\% of genuine pairs, whereas Ovis1.5 reaches only 34.6\%. At the stricter FMR\,=\,0.1\% threshold, only four models produce scores with sufficient granularity to reach this operating point.

\textbf{RFW.} All models perform worse on RFW than on IJB-C. FaceLLM-8B remains the best-performing model, but its EER rises from 5.13\% to 29.46\%; Qwen2-VL-7B similarly increases from 8.54\% to 34.39\%. Several models that are competitive on IJB-C, such as Idefics3-8B (35.08\%) and Qwen2.5-VL-7B (34.98\%), produce similar EER values around 35\% on RFW. Ovis1.5 and Qwen2-VL-2B approach 50\% EER, which corresponds to random guessing.

This difference in model performance across these two datasets can be explained by the number of images per template: on average, IJB-C templates contain 20.3 media items (median 9), whereas RFW templates contain 3.6 (median 3). Since MLLM-based verification relies on averaging pairwise scores across all media items in a template comparison, the larger templates of IJB-C allow for better performance compared to RFW as the MLLMs have information to process per identity.

\begin{figure*}[!htb]
  \centering
  \scalebox{0.825} {
  \includegraphics[width=\linewidth]{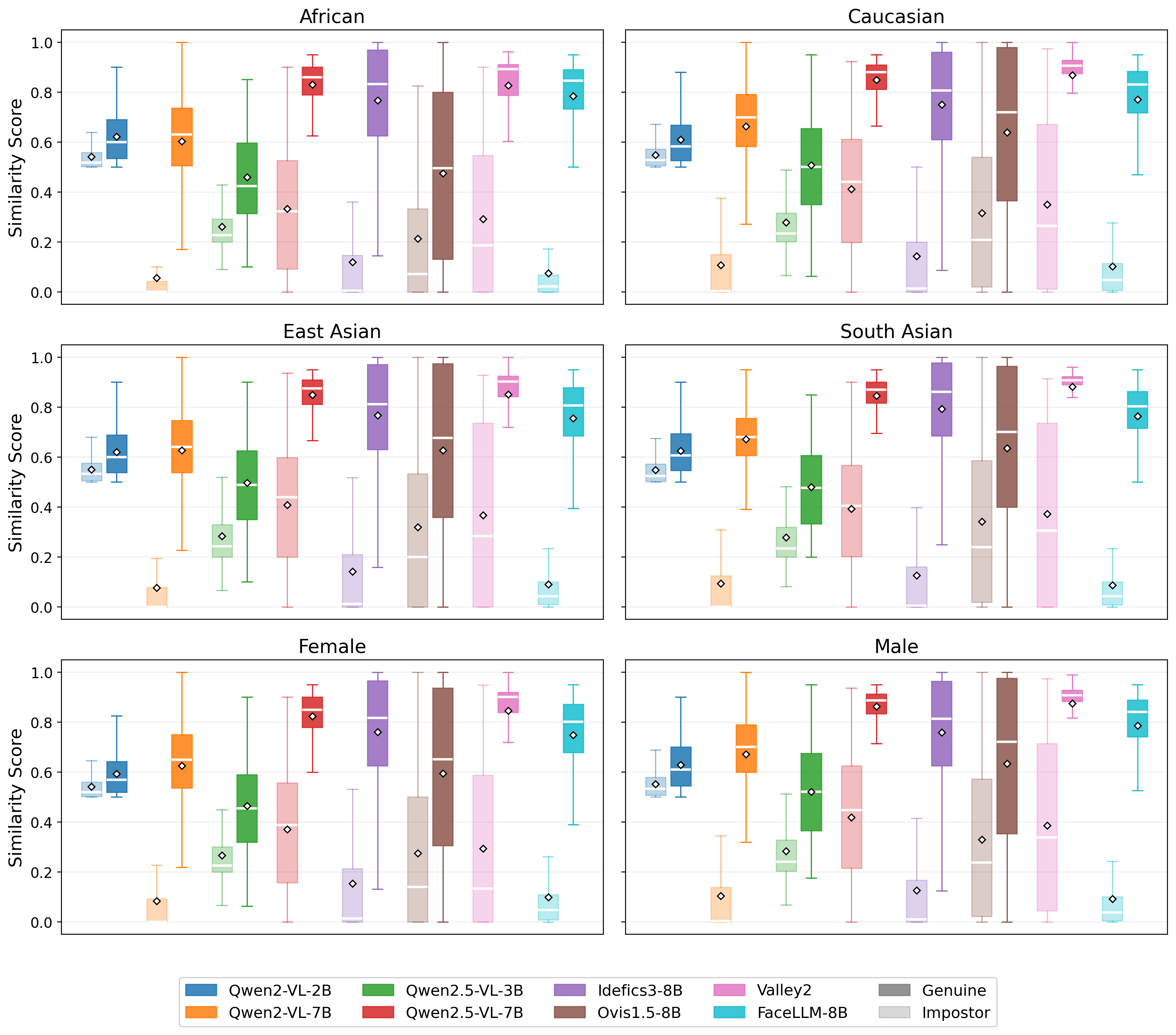}
   }
  \caption{Genuine (dark) and impostor (light) score distributions on IJB-C, stratified by ethnicity (top four panels) and gender (bottom two panels).}
  
  \label{fig:score_boxplot}
\end{figure*}

\Cref{fig:score_boxplot} shows the genuine and impostor score distributions for each demographic group. FaceLLM-8B shows the clearest separation between the two distributions on both benchmarks, which is consistent with its low EER. Ovis1.5 and Qwen2-VL-2B, on the other hand, have heavily overlapping genuine and impostor distributions, which explains their near-chance accuracy. The $d'$ (\cref{eq:dprime}), reported in the last column of \cref{tab:fairness}, provides a numerical measure of this separation. FaceLLM-8B has the highest mean $d'$ on IJB-C (4.50) and on RFW (1.08), while the weakest models fall below $d' = 1$.

% Figure: Score distribution boxplots — 1x6 grid (4 ethnicity + 2 gender)

\begin{figure*}[!htb]
  \centering
  \includegraphics[width=\linewidth]{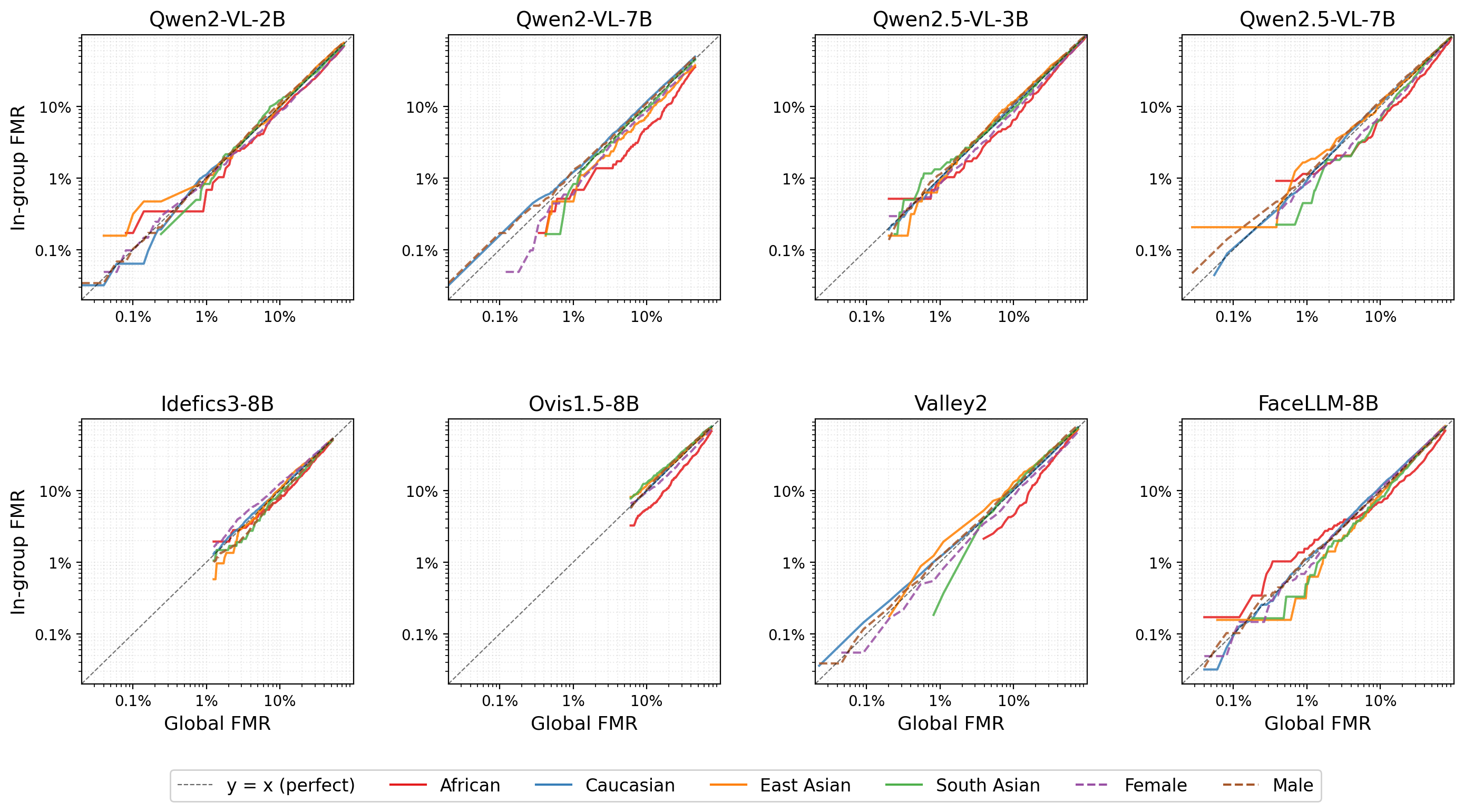}
  \caption{Global FMR vs.\ in-group FMR at every operating threshold for each model on IJB-C.  Solid lines show ethnicity groups; dashed lines show gender groups.  Points below the diagonal indicate privilege for that group as those ones achieve less-than-average error rates.}
  \label{fig:global_ingroup}
\end{figure*}

\subsection{Bias and Fairness Evaluation}
\label{sec:efairness}

\Cref{tab:fairness} reports four FMR-based fairness metrics evaluated at the EER threshold and at three fixed operating points (FMR\,=\,10\%, 1\%, 0.1\%), together with the mean decidability index $d'$.
The metrics are the FMR differential $\Delta_\text{F}$ (maximum minus minimum per-group FMR), the FMR ratio $R_\text{F}$ (maximum divided by minimum per-group FMR), the multi-group metric $M_\text{F}$ (maximum per-group FMR divided by the geometric mean), and the Gini coefficient $G_\text{F}$ (overall inequality of per-group FMR values). \Cref{fig:global_ingroup} plots in-group FMR against the global FMR across the full range of decision thresholds for both gender and ethnicity on the IJB-C dataset. 

\textbf{Ethnicity fairness in IJB-C.}
% note to self: simplify this, it's too dense
As shown in \cref{tab:performance}, the Caucasian group has the highest per-group EER in five of the eight non-degenerate models. This is a notable finding, since in many embedding-based face recognition systems, darker-skinned groups tend to be the most disadvantaged~\cite{frvt-3, gendershades, jjhoward-1}.
At the EER threshold, FaceLLM-8B has the lowest differential ($\Delta_\text{F} = 0.020$), yet its Gini coefficient ($G_\text{F} = 0.092$) is higher than those of Qwen2-VL-2B ($G_\text{F} = 0.059$) and Idefics3-8B ($G_\text{F} = 0.064$). This difference can be explained by the fact that $\Delta_\text{F}$ is an absolute measure that scales with overall FMR, whereas $G_\text{F}$ captures relative inequality among the groups. At stricter operating points, $R_\text{F}$ and $M_\text{F}$ increase sharply and often become undefined, because per-group FMR values approach zero at different rates. When this happens, small absolute differences ($\Delta_\text{F}$) turn into very large ratios, and the ratio-based metrics lose their interpretability for many model-threshold combinations. The Gini coefficient also increases at stricter thresholds, indicating that relative FMR inequality grows even when absolute gaps remain small.

\textbf{Ethnicity fairness in RFW.}
The RFW block of \cref{tab:fairness} shows larger ethnicity disparities than IJB-C at the EER threshold.
FaceLLM-8B, which has the best overall accuracy on RFW, also has the highest differential ($\Delta_\text{F} = 0.478$) and one of the highest Gini values ($G_\text{F} = 0.293$) among the eight models. By contrast, Ovis1.5 has the lowest Gini ($G_\text{F} = 0.020$) but also performs near chance level (49.18\% EER). This time, the African group is the most disadvantaged across all eight models. E.g., FaceLLM-8B has a 21.23\% EER on the Asian group but 35.25\% on the African group, a gap of 14 percentage points. The per-ethnicity EER standard deviations ($\sigma$) in \cref{tab:performance} confirm this pattern, with values between 1.77 and 5.58 on RFW compared to 0.56--2.75 on IJB-C.

\textbf{Gender fairness in IJB-C.}
The gender block of \cref{tab:fairness} shows smaller disparities than the corresponding ethnicity block across all four metrics and all operating points. At the EER threshold, FaceLLM-8B has very low gender disparity ($\Delta_\text{F} = 0.001$, $G_\text{F} = 0.004$). Even at FMR\,=\,10\%, the maximum gender Gini across all models is 0.123 (Qwen2.5-VL-7B), compared to a maximum ethnicity Gini of 0.172 (Valley2). The per-gender EER differences ($\Delta$ in \cref{tab:performance}) support this observation. Qwen2-VL-7B has the smallest absolute gender gap ($\Delta = 0.10$ pp), while Qwen2-VL-2B has the largest ($\Delta = 3.77$ pp). In the global-vs-ingroup plots of \cref{fig:global_ingroup}, the dashed gender curves lie closer to the diagonal than the solid ethnicity curves in nearly every model panel, which indicates that gender has a smaller effect on false match rates than ethnicity.

\subsection{Accuracy-Fairness Trade-off}
\label{sec:tradeoff}

The results in \cref{tab:fairness,tab:performance} show that the most accurate models are not necessarily the fairest, and that the fairest models are often the least accurate. At FMR\,=\,10\% on IJB-C ethnicity, the two lowest Gini values belong to Qwen2-VL-2B ($G_\text{F} = 0.068$) and Idefics3-8B ($G_\text{F} = 0.070$), which rank seventh and fourth in overall accuracy, respectively. FaceLLM-8B, the most accurate model, ranks only third in terms of fairness ($G_\text{F} = 0.099$). On RFW, this pattern is more prevalent. Ovis1.5 has the lowest ethnicity Gini at the EER threshold ($G_\text{F} = 0.020$) while also having the second-worst EER (49.18\%). The reason is that models with poor ability to separate genuine and impostor scores produce similar false match rates across all demographic groups, and therefore appear ``fair'' despite being practically sub-optimal. On the other hand, FaceLLM-8B combines the best accuracy with the lowest gender disparity on IJB-C ($G_\text{F} = 0.004$ at EER, $d' = 4.31$), which shows that accuracy and fairness are not necessarily opposed when the score distributions are well separated.

\section{Conclusion}
\label{sec:conclusion}

This paper presented the first demographic fairness evaluation of MLLMs for face verification. We evaluated nine MLLMs from six model families, ranging from 2B to 8B parameters, on the RFW and (a sub-sample of) IJB-C 1:1 verification protocols with respect to four ethnicity groups and two gender groups. We reported four FMR-based fairness metrics alongside standard accuracy measures at multiple operating points. The results indicate that current MLLMs still fall short of the accuracy and speed needed for practical face verification. FaceLLM-8B, the only face-specialised model in our study, outperforms all general-purpose models on both benchmarks. The fairness analysis produced several findings that differ from common patterns in embedding-based face recognition: On IJB-C, the Caucasian group is the most disadvantaged in five of eight models, which is the opposite of what is typically reported for traditional face recognition systems. Gender disparities are consistently smaller than ethnicity disparities across all models and operating points, with FaceLLM-8B showing low gender inequality ($G_\text{F} = 0.004$ at the EER threshold).

\paragraph{Limitations and future work.}
Several directions are worth exploring. The evaluation could be extended to additional benchmarks and intersectional attributes (ethnicity~$\times$~gender) to provide a more complete view of MLLM bias. The sensitivity of verification scores to prompt phrasing and in-context examples should be studied, since the current results depend on a single prompt template. Evaluating larger MLLMs ($>$13B parameters) and instruction-tuned variants would help determine whether increasing model scale can close the accuracy gap with embedding-based systems.

\section{Acknowledgments}
This work has received funding from the European Union's Horizon Europe research and innovation programme under grant agreement No.~101189650 (CERTAIN: \emph{Certification for Ethical and Regulatory Transparency in Artificial Intelligence}). 
This work was also funded by the European Union project CarMen (Grant Agreement No. 101168325).

%%% Commented-out template sections (kept for reference):
% \input{sec/2_formatting}
% \input{sec/3_finalcopy}

{
    \small
    \bibliographystyle{ieeenat_fullname}
    \bibliography{main}

@String(CVPR= {IEEE Conf. Comput. Vis. Pattern Recog.})

@String(ICCV= {Int. Conf. Comput. Vis.})

@String(ICPR = {Int. Conf. Pattern Recog.})

@String(ICIP = {IEEE Int. Conf. Image Process.})

@String(CVPR  = {CVPR})

@String(ICCV  = {ICCV})

@String(ICPR  = {ICPR})

@String(ICIP  = {ICIP})

@article{narayan2025facexbench,
  title={FaceXBench: Evaluating Multimodal LLMs on Face Understanding},
  author={Narayan, Kartik and VS, Vibashan and Patel, Vishal M},
  journal={arXiv preprint arXiv:2501.10360},
  year={2025}
}

@inproceedings{jia2024can,
  title={Can chatgpt detect deepfakes? a study of using multimodal large language models for media forensics},
  author={Jia, Shan and Lyu, Reilin and Zhao, Kangran and Chen, Yize and Yan, Zhiyuan and Ju, Yan and Hu, Chuanbo and Li, Xin and Wu, Baoyuan and Lyu, Siwei},
  booktitle={Proceedings of the IEEE/CVF Conference on Computer Vision and Pattern Recognition},
  pages={4324--4333},
  year={2024}
}

@inproceedings{hassanpour2024chatgpt,
  title={ChatGPT and biometrics: an assessment of face recognition, gender detection, and age estimation capabilities},
  author={Hassanpour, Ahmad and Kowsari, Yasamin and Shahreza, Hatef Otroshi and Yang, Bian and Marcel, S{\'e}bastien},
  booktitle={2024 IEEE International Conference on Image Processing (ICIP)},
  pages={3224--3229},
  year={2024},
  organization={IEEE}
}

@article{komaty2025exploring,
  title={Exploring ChatGPT for Face Presentation Attack Detection in Zero and Few-Shot in-Context Learning},
  author={Komaty, Alain and Shahreza, Hatef Otroshi and George, Anjith and Marcel, Sebastien},
  journal={arXiv preprint arXiv:2501.08799},
  year={2025}
}

@article{shahreza2025foundation,
  title={Foundation models and biometrics: A survey and outlook},
  author={Shahreza, Hatef Otroshi and Marcel, S{\'e}bastien},
  journal={IEEE Transactions on Information Forensics and Security},
  year={2025},
  publisher={IEEE}
}

@article{shi2024shield,
  title={Shield: An evaluation benchmark for face spoofing and forgery detection with multimodal large language models},
  author={Shi, Yichen and others},
  journal={arXiv preprint arXiv:2402.04178},
  year={2024}
}

@article{hurst2024gpt,
  title={GPT-4o System Card},
  author={Hurst, Aaron and Lerer, Adam and Goucher, Adam P and Perelman, Adam and Ramesh, Aditya and Clark, Aidan and Ostrow, AJ and Welihinda, Akila and Hayes, Alan and Radford, Alec and others},
  journal={arXiv preprint arXiv:2410.21276},
  year={2024}
}

@article{team2023gemini,
  title={Gemini: a family of highly capable multimodal models},
  author={Team, Gemini and Anil, Rohan and Borgeaud, Sebastian and Alayrac, Jean-Baptiste and Yu, Jiahui and Soricut, Radu and Schalkwyk, Johan and Dai, Andrew M and Hauth, Anja and Millican, Katie and others},
  journal={arXiv preprint arXiv:2312.11805},
  year={2023}
}

@article{wang2025facebench,
  title={FaceBench: A Multi-View Multi-Level Facial Attribute VQA Dataset for Benchmarking Face Perception MLLMs},
  author={Wang, Xiaoqin and Ma, Xusen and Hou, Xianxu and Ding, Meidan and Li, Yudong and Chen, Junliang and Chen, Wenting and Peng, Xiaoyang and Shen, Linlin},
  journal={arXiv preprint arXiv:2503.21457},
  year={2025}
}

@article{facerecbench2025,
    author    = {Hatef Otroshi Shahreza and S{\'e}bastien Marcel},
    title     = {Benchmarking Multimodal Large Language Models for Face Recognition},
    journal   = {arXiv preprint arXiv:2510.14866},
    year      = {2025}
  }

@misc{laurencon2024building,
      title={Building and better understanding vision-language models: insights and future directions.}, 
      author={Hugo Laurençon and Andrés Marafioti and Victor Sanh and Léo Tronchon},
      year={2024},
      eprint={2408.12637},
      archivePrefix={arXiv},
      primaryClass={cs.CV}
}

@article{lu2024ovis,
  title={Ovis: Structural Embedding Alignment for Multimodal Large Language Model}, 
  author={Shiyin Lu and Yang Li and Qing-Guo Chen and Zhao Xu and Weihua Luo and Kaifu Zhang and Han-Jia Ye},
  year={2024},
  journal={arXiv:2405.20797}
}

@article{Qwen2VL,
  title={Qwen2-VL: Enhancing Vision-Language Model's Perception of the World at Any Resolution},
  author={Wang, Peng and Bai, Shuai and Tan, Sinan and Wang, Shijie and Fan, Zhihao and Bai, Jinze and Chen, Keqin and Liu, Xuejing and Wang, Jialin and Ge, Wenbin and Fan, Yang and Dang, Kai and Du, Mengfei and Ren, Xuancheng and Men, Rui and Liu, Dayiheng and Zhou, Chang and Zhou, Jingren and Lin, Junyang},
  journal={arXiv preprint arXiv:2409.12191},
  year={2024}
}

@misc{qwen2025qwen25technicalreport,
      title={Qwen2.5 Technical Report}, 
      author={Qwen and : and An Yang and Baosong Yang and Beichen Zhang and Binyuan Hui and Bo Zheng and Bowen Yu and Chengyuan Li and Dayiheng Liu and Fei Huang and Haoran Wei and others},
      year={2025},
      eprint={2412.15115},
      archivePrefix={arXiv},
      primaryClass={cs.CL},
      url={https://arxiv.org/abs/2412.15115}, 
}

@article{wu2025valley2,
  title={Valley2: Exploring Multimodal Models with Scalable Vision-Language Design},
  author={Wu, Ziheng and Chen, Zhenghao and Luo, Ruipu and Zhang, Can and Gao, Yuan and He, Zhentao and Wang, Xian and Lin, Haoran and Qiu, Minghui},
  journal={arXiv preprint arXiv:2501.05901},
  year={2025}
}

@misc{liu2023improvedllava,
      author={Liu, Haotian and Li, Chunyuan and Li, Yuheng and Lee, Yong Jae},
      title={Improved Baselines with Visual Instruction Tuning}, 
      publisher={arXiv:2310.03744},
      year={2023},
}

@article{shahreza2025facellm,
  title={FaceLLM: A Multimodal Large Language Model for Face Understanding},
  author={Shahreza, Hatef Otroshi and Marcel, S{\'e}bastien},
  journal={arXiv preprint arXiv:2507.10300},
  year={2025}
}

@InProceedings{gendershades,
  title = 	 {Gender Shades: Intersectional Accuracy Disparities in Commercial Gender Classification},
  author = 	 {Buolamwini, Joy and Gebru, Timnit},
  booktitle = 	 {Proceedings of the 1st Conference on Fairness, Accountability and Transparency},
  pages = 	 {77--91},
  year = 	 {2018},
  editor = 	 {Friedler, Sorelle A. and Wilson, Christo},
  volume = 	 {81},
  series = 	 {Proceedings of Machine Learning Research},
  month = 	 {23--24 Feb},
  publisher =    {PMLR},
  url = 	 {https://proceedings.mlr.press/v81/buolamwini18a.html}
}

@misc{frvt-3,
  author = {Patrick Grother and Mei Ngan and Kayee Hanaoka},
  title = {Face recognition vendor test part 3::demographic effects},
  year = {2019},
  month = {2019-12-01 05:12:00},
  publisher = {, National Institute of Standards and Technology, Gaithersburg, MD},
  doi = {https://doi.org/10.6028/NIST.IR.8280},
  language = {en},
}

@misc{frvt-8,
  author = {Patrick Grother},
  title = {Face recognition vendor test (FRVT) part 8::summarizing demographic differentials},
  year = {2022},
  month = {2022-07-01 00:00:00},
  publisher = {, National Institute of Standards and Technology, Gaithersburg, MD},
  doi = {https://doi.org/10.6028/NIST.IR.8429.ipd},
  language = {en},
}

@INPROCEEDINGS{jjhoward-1,
  author={Howard, John J. and Sirotin, Yevgeniy B. and Vemury, Arun R.},
  booktitle={2019 IEEE 10th International Conference on Biometrics Theory, Applications and Systems (BTAS)}, 
  title={The Effect of Broad and Specific Demographic Homogeneity on the Imposter Distributions and False Match Rates in Face Recognition Algorithm Performance}, 
  year={2019},
  volume={},
  number={},
  pages={1-8},
  doi={10.1109/BTAS46853.2019.9186002}}

@inproceedings{ketan-fairness-distribution,
author = {Kotwal, Ketan and Marcel, S\'{e}bastien},
title = {Fairness Index Measures to Evaluate Bias in Biometric Recognition},
year = {2022},
isbn = {978-3-031-37659-7},
publisher = {Springer-Verlag},
address = {Berlin, Heidelberg},
url = {https://doi.org/10.1007/978-3-031-37660-3_34},
doi = {10.1007/978-3-031-37660-3_34},
booktitle = {Pattern Recognition, Computer Vision, and Image Processing. ICPR 2022 International Workshops and Challenges: Montreal, QC, Canada, August 21–25, 2022, Proceedings, Part I},
pages = {479–493},
numpages = {15},
location = {Montr\'{e}al, QC, Canada}
}

@article{ketan-fairness-survey,
   title={Review of Demographic Fairness in Face Recognition},
   volume={8},
   ISSN={2637-6407},
   url={http://dx.doi.org/10.1109/TBIOM.2025.3601217},
   DOI={10.1109/tbiom.2025.3601217},
   number={1},
   journal={IEEE Transactions on Biometrics, Behavior, and Identity Science},
   publisher={Institute of Electrical and Electronics Engineers (IEEE)},
   author={Kotwal, Ketan and Marcel, Sébastien},
   year={2026},
   month=jan, pages={20–45} }

@InProceedings{garbe,
author="Howard, John J.
and Laird, Eli J.
and Rubin, Rebecca E.
and Sirotin, Yevgeniy B.
and Tipton, Jerry L.
and Vemury, Arun R.",
editor="Rousseau, Jean-Jacques
and Kapralos, Bill",
title="Evaluating Proposed Fairness Models for Face Recognition Algorithms",
booktitle="Pattern Recognition, Computer Vision, and Image Processing. ICPR 2022 International Workshops and Challenges",
year="2023",
publisher="Springer Nature Switzerland",
address="Cham",
pages="431--447",
isbn="978-3-031-37660-3"
}

@ARTICLE{fdr,
  author={de Freitas Pereira, Tiago and Marcel, Sébastien},
  journal={IEEE Transactions on Biometrics, Behavior, and Identity Science}, 
  title={Fairness in Biometrics: A Figure of Merit to Assess Biometric Verification Systems}, 
  year={2022},
  volume={4},
  number={1},
  pages={19-29},
  doi={10.1109/TBIOM.2021.3102862}}

@INPROCEEDINGS{ijbc,
  author={Maze, Brianna and Adams, Jocelyn and Duncan, James A. and Kalka, Nathan and Miller, Tim and Otto, Charles and Jain, Anil K. and Niggel, W. Tyler and Anderson, Janet and Cheney, Jordan and Grother, Patrick},
  booktitle={2018 International Conference on Biometrics (ICB)}, 
  title={IARPA Janus Benchmark - C: Face Dataset and Protocol}, 
  year={2018},
  volume={},
  number={},
  pages={158-165},
  doi={10.1109/ICB2018.2018.00033}}

@INPROCEEDINGS{rfw,
  author={Wang, Mei and Deng, Weihong and Hu, Jiani and Tao, Xunqiang and Huang, Yaohai},
  booktitle={2019 IEEE/CVF International Conference on Computer Vision (ICCV)}, 
  title={Racial Faces in the Wild: Reducing Racial Bias by Information Maximization Adaptation Network}, 
  year={2019},
  volume={},
  number={},
  pages={692-702},
  doi={10.1109/ICCV.2019.00078}}

@INPROCEEDINGS{hairy,
  author={Bhatta, Aman and Albiero, Vítor and Bowyer, Kevin W. and King, Michael C.},
  booktitle={2023 IEEE/CVF Winter Conference on Applications of Computer Vision Workshops (WACVW)}, 
  title={The Gender Gap in Face Recognition Accuracy Is a Hairy Problem}, 
  year={2023},
  volume={},
  number={},
  pages={1-10},
  doi={10.1109/WACVW58289.2023.00034}}

@INPROCEEDINGS{face-regions,
  author={Albiero, Vítor and Bowyer, Kevin W. and King, Michael C.},
  booktitle={2022 IEEE International Joint Conference on Biometrics (IJCB)}, 
  title={Face Regions Impact Recognition Accuracy Differently Across Demographics}, 
  year={2022},
  volume={},
  number={},
  pages={1-9},
  doi={10.1109/IJCB54206.2022.10007941}}

@InProceedings{demographic-bias-analysis,
author="Sarridis, Ioannis
and Koutlis, Christos
and Papadopoulos, Symeon
and Diou, Christos",
editor="Meo, Rosa
and Silvestri, Fabrizio",
title="Towards Fair Face Verification: An In-depth Analysis of Demographic Biases",
booktitle="Machine Learning and Principles and Practice of Knowledge Discovery in Databases",
year="2025",
publisher="Springer Nature Switzerland",
address="Cham",
pages="194--208",
isbn="978-3-031-74630-7"
}

@techreport{iso19795-10:2024,
  type        = {Standard},
  key         = {ISO/IEC 19795-10:2024},
  institution = {International Organization for Standardization},
  title       = {{Information technology — Biometric performance testing and reporting — Part 10: Quantifying biometric system performance variation across demographic groups}},
  number      = {ISO/IEC 19795-10:2024},
  year        = {2024},
  address     = {Geneva, Switzerland},
  url         = {https://www.iso.org/standard/81223.html}
}

@InProceedings{facenet,
author = {Schroff, Florian and Kalenichenko, Dmitry and Philbin, James},
title = {FaceNet: A Unified Embedding for Face Recognition and Clustering},
booktitle = {Proceedings of the IEEE Conference on Computer Vision and Pattern Recognition (CVPR)},
month = {June},
year = {2015}
}

@InProceedings{arcface,
author = {Deng, Jiankang and Guo, Jia and Xue, Niannan and Zafeiriou, Stefanos},
title = {ArcFace: Additive Angular Margin Loss for Deep Face Recognition},
booktitle = {Proceedings of the IEEE/CVF Conference on Computer Vision and Pattern Recognition (CVPR)},
month = {June},
year = {2019}
}

@inproceedings{cosface,
  author       = {Hao Wang and
                  Yitong Wang and
                  Zheng Zhou and
                  Xing Ji and
                  Dihong Gong and
                  Jingchao Zhou and
                  Zhifeng Li and
                  Wei Liu},
  title        = {CosFace: Large Margin Cosine Loss for Deep Face Recognition},
  booktitle    = {2018 {IEEE} Conference on Computer Vision and Pattern Recognition,
                  {CVPR} 2018, Salt Lake City, UT, USA, June 18-22, 2018},
  pages        = {5265--5274},
  publisher    = {Computer Vision Foundation / {IEEE} Computer Society},
  year         = {2018},
  url          = {http://openaccess.thecvf.com/content\_cvpr\_2018/html/Wang\_CosFace\_Large\_Margin\_CVPR\_2018\_paper.html},
  doi          = {10.1109/CVPR.2018.00552},
}

@INPROCEEDINGS{adaface,
  author={Kim, Minchul and Jain, Anil K. and Liu, Xiaoming},
  booktitle={2022 IEEE/CVF Conference on Computer Vision and Pattern Recognition (CVPR)}, 
  title={AdaFace: Quality Adaptive Margin for Face Recognition}, 
  year={2022},
  volume={},
  number={},
  pages={18729-18738},
  doi={10.1109/CVPR52688.2022.01819}}

@INPROCEEDINGS{magface,
  author={Meng, Qiang and Zhao, Shichao and Huang, Zhida and Zhou, Feng},
  booktitle={2021 IEEE/CVF Conference on Computer Vision and Pattern Recognition (CVPR)}, 
  title={MagFace: A Universal Representation for Face Recognition and Quality Assessment}, 
  year={2021},
  volume={},
  number={},
  pages={14220-14229},
  doi={10.1109/CVPR46437.2021.01400}}

@article{mllm-survey,
    author = {Yin, Shukang and Fu, Chaoyou and Zhao, Sirui and Li, Ke and Sun, Xing and Xu, Tong and Chen, Enhong},
    title = {A survey on multimodal large language models},
    journal = {National Science Review},
    volume = {11},
    number = {12},
    pages = {nwae403},
    year = {2024},
    month = {11},
    issn = {2095-5138},
    doi = {10.1093/nsr/nwae403},
    url = {https://doi.org/10.1093/nsr/nwae403},
    eprint = {https://academic.oup.com/nsr/article-pdf/11/12/nwae403/61201557/nwae403.pdf},
}

@inproceedings{llava,
 author = {Liu, Haotian and Li, Chunyuan and Wu, Qingyang and Lee, Yong Jae},
 booktitle = {Advances in Neural Information Processing Systems},
 editor = {A. Oh and T. Naumann and A. Globerson and K. Saenko and M. Hardt and S. Levine},
 pages = {34892--34916},
 publisher = {Curran Associates, Inc.},
 title = {Visual Instruction Tuning},
 url = {https://proceedings.neurips.cc/paper_files/paper/2023/file/6dcf277ea32ce3288914faf369fe6de0-Paper-Conference.pdf},
 volume = {36},
 year = {2023}
}

@article{llama,
  author       = {Hugo Touvron and
                  Thibaut Lavril and
                  Gautier Izacard and
                  Xavier Martinet and
                  Marie{-}Anne Lachaux and
                  Timoth{\'{e}}e Lacroix and
                  Baptiste Rozi{\`{e}}re and
                  Naman Goyal and
                  Eric Hambro and
                  Faisal Azhar and
                  Aur{\'{e}}lien Rodriguez and
                  Armand Joulin and
                  Edouard Grave and
                  Guillaume Lample},
  title        = {LLaMA: Open and Efficient Foundation Language Models},
  journal      = {CoRR},
  volume       = {abs/2302.13971},
  year         = {2023},
  url          = {https://doi.org/10.48550/arXiv.2302.13971},
  doi          = {10.48550/ARXIV.2302.13971},
  eprinttype    = {arXiv},
  eprint       = {2302.13971}
}

@article{google-emergent,
    title	= {Emergent abilities of large language models},
    author	= {Barret Zoph and Colin Raffel and Dale Schuurmans and Dani Yogatama and Denny Zhou and Don Metzler and Ed H. Chi and Jason Wei and Jeff Dean and Liam B. Fedus and Maarten Paul Bosma and Oriol Vinyals and Percy Liang and Sebastian Borgeaud and Tatsunori B. Hashimoto and Yi Tay},
    year	= {2022},
    journal	= {TMLR}
    }

@misc{claude35,
    title={Claude 3.5 Sonnet Model Card Addendum},
    author={{Anthropic}},
    year={2024},
    url={https://www-cdn.anthropic.com/fed9cc193a14b84131812372d8d5857f8f304c52/Model_Card_Claude_3_Addendum.pdf}
}

@ARTICLE{drozdowski-bias-survey,
  author={Drozdowski, Pawel and Rathgeb, Christian and Dantcheva, Antitza and Damer, Naser and Busch, Christoph},
  journal={IEEE Transactions on Technology and Society}, 
  title={Demographic Bias in Biometrics: A Survey on an Emerging Challenge}, 
  year={2020},
  volume={1},
  number={2},
  pages={89-103},
  doi={10.1109/TTS.2020.2992344}}

@article{shahreza2026evaluating,
  title={Evaluating Multimodal Large Language Models for Heterogeneous Face Recognition},
  author={Shahreza, Hatef Otroshi and George, Anjith and Marcel, S{\'e}bastien},
  journal={arXiv preprint arXiv:2601.15406},
  year={2026}
}

@ARTICLE{decidability,
  author={Daugman, J.},
  journal={IEEE Transactions on Circuits and Systems for Video Technology}, 
  title={How iris recognition works}, 
  year={2004},
  volume={14},
  number={1},
  pages={21-30},
  doi={10.1109/TCSVT.2003.818350}}
}


\end{document}